\documentclass[letterpaper, 10 pt, conference]{ieeeconf}  

\IEEEoverridecommandlockouts  

\let\labelindent\relax
\usepackage[left=1.88cm,right=1.88cm,top=1.88cm,bottom=1.88cm]{geometry}
\usepackage{amsmath,amssymb,amsfonts}
\usepackage{textcomp}
\usepackage{xcolor}
\usepackage{bbm}
\usepackage{enumitem}
\usepackage{tabularx, booktabs}
\usepackage{soul,color}
\usepackage{siunitx}
\usepackage{dsfont}
\usepackage{multirow}

\usepackage{times}
\usepackage{multicol}
\usepackage{hyperref}
\usepackage{cleveref}
\usepackage{bm}
\usepackage[nolist]{acronym}
\usepackage[noend]{algpseudocode}
\usepackage[linesnumbered,ruled]{algorithm2e}

\usepackage{mathtools}
\usepackage{array}
\usepackage{dblfloatfix}
\usepackage{subcaption}
\usepackage{caption}
\usepackage[font=small]{caption}
\usepackage{subcaption}
\usepackage{graphicx}
\usepackage{wrapfig}
\usepackage{diagbox}
\usepackage{etoolbox}\AtBeginEnvironment{algorithmic}{\small}
\usepackage{color,soul} 

\newcommand\clearrow{\global\let\rowmac\relax}
\usepackage{lipsum}
\usepackage{color}
\usepackage[colorinlistoftodos]{todonotes}

\newcommand{\vect}[1]{\boldsymbol{#1}}

\newtheorem{definition}{Definition}
\newtheorem{problem}{Problem}
\newtheorem{remark}{Remark}

\usepackage{bbm} 
\usepackage{array}

\newcolumntype{C}[1]{>{\centering\let\newline\\\arraybackslash\hspace{0pt}}m{#1}}

\newcolumntype{P}[1]{>{\centering\arraybackslash}p{#1}}

\title{\LARGE \bf
Graph-based Path Planning with Dynamic Obstacle Avoidance for Autonomous Parking
}

\author{Farhad Nawaz$^{*1, 2}$, Minjun Sung$^{1, 3}$, Darshan Gadginmath$^{1, 4}$, Jovin D'sa$^{1}$, Sangjae Bae$^{1}$, \\ David Isele$^{1}$, Nadia Figueroa$^{2}$, Nikolai Matni$^{2}$ and Faizan M. Tariq$^{*1}$
\thanks{
$^{1}$Honda Research Institute (HRI), San Jose, CA 95134, USA.}
\thanks{$^{2}$GRASP Lab, University of Pennsylvania, PA 19104, USA.}
\thanks{$^{3}$University of Illinois at Urbana-Champaign, USA.}
\thanks{$^{4}$University of California at Riverside, Riverside, CA, 92521, USA}
\thanks{$^{*}$Corresponding authors: {\tt farhadn@seas.upenn.edu} \& \newline {\tt faizan\_tariq@honda-ri.com}}
\thanks{
All work was done when Farhad Nawaz, Minjun Sung and Darshan Gadginmath were employed by HRI.}
}

\begin{document}

\maketitle
\thispagestyle{empty}
\pagestyle{empty}

\begin{abstract}

Safe and efficient path planning in parking scenarios presents a significant challenge due to the presence of cluttered environments filled with static and dynamic obstacles. To address this, we propose a novel and computationally efficient planning strategy that seamlessly integrates the predictions of dynamic obstacles into the planning process, ensuring the generation of collision-free paths. Our approach builds upon the conventional Hybrid A star algorithm by introducing a time-indexed variant that explicitly accounts for the predictions of dynamic obstacles during node exploration in the graph, thus enabling dynamic obstacle avoidance. We integrate the \textit{time-indexed} Hybrid A star algorithm within an online planning framework to compute local paths at each planning step, guided by an adaptively chosen intermediate goal. The proposed method is validated in diverse parking scenarios, including perpendicular, angled, and parallel parking. Through simulations, we showcase our approach's potential in greatly improving the efficiency and safety when compared to the state of the art spline-based planning method for parking situations.
\end{abstract}

\vspace{-15pt}
\section{INTRODUCTION}

(Semi)autonomous parking addresses the growing demand for efficient and safe vehicle maneuvering in constrained environments~\cite{valet_pappas, urban_autonomous}. Urbanization and increased vehicle traffic have resulted in congested parking lots, necessitating precise, collision-free navigation. This task is further complicated by the presence of dynamic obstacles~\cite{dynamic_parking}, such as pedestrians and moving vehicles, in addition to static obstacles~\cite{static_parking}, including parked vehicles and structural elements. A functional planning module for autonomous parking must account for these complexities while ensuring safety, reliability, and computational efficiency. 

\begin{figure}[!t]
         \centering         \includegraphics[width=0.85\linewidth]{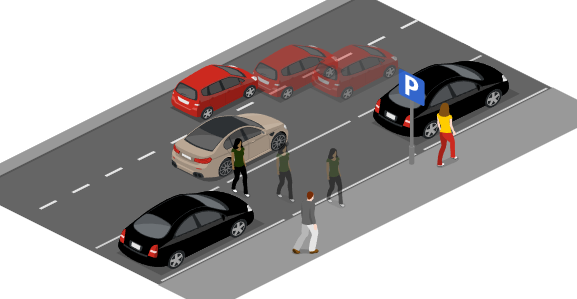}
      \caption{A parallel parking scenario with static cars (black), pedestrians, and a moving car (red). The brown car is the ego vehicle. The more transparent images of the red car and the pedestrian denote their respective predictions.} 
       \vspace{-20pt}
        \label{fig:intro}
\end{figure}

Consider the scenario in Fig.~\ref{fig:intro}, where the brown ego vehicle is trying to park in the spot between the two black vehicles. Meanwhile, a red car in the adjacent lane is overtaking the ego vehicle, and a pedestrian is walking across the empty parking spot to get on the sidewalk. The more transparent images denoted the predictions of the pedestrian and the red car. Path planning techniques in such scenarios must effectively model and avoid both static and dynamic obstacles, generate precise maneuvers for navigating tight spaces, and ensure computational efficiency to react swiftly in dynamic environments. 
The complexity is primarily attributed to the non-convex geometry of the obstacle-free space, in addition to the non-linear and non-holonomic nature of vehicle dynamics~\cite{rajamani2011vehicle}, which impose strict motion constraints during the planning process. Furthermore, it has been formally established that identifying a collision-free path in such scenarios is, in general, an NP-hard problem~\cite{NP_hard}, underscoring the computational intractability of this task. Thus, the current research gap is to generate path that simultaneously ensure (i) real-time efficiency, (ii) kinematic feasibility, and (iii) safety in the presence of both static and dynamic obstacles. To address these problems, \textit{we develop a computationally efficient planning strategy that generates safe and reliable paths for autonomous parking maneuvers by explicitly accounting for the motion of dynamic obstacles in our graph-based search algorithm.} We summarize our contributions below. 

\textbf{Contributions:} We propose a novel \textit{time-indexed variant} of the conventional Hybrid A$^\star$ algorithm that explicitly uses the \textit{motion predictions of dynamic obstacles} to generate collision-free paths. Then, we present a strategy for path planning in larger parking lots that utilizes the \textit{time-indexed} Hybrid A$^\star$ algorithm as a sub-routine to compute local paths at each planning step by choosing an adaptive intermediate goal based on look-ahead point from the current state. We exploit the static map information and incorporate the vanilla A* cost as a heuristic to guide the ego vehicle towards the goal, resulting in improved computational performance and dynamically feasible paths.

We demonstrate through simulations in diverse parking scenarios that our method is computationally efficient compared to the state of the art spline-based approach while generating safer and smoother paths. The simulation videos are available at~\href{https://sites.google.com/view/t-ha-star/home}{https://sites.google.com/view/t-ha-star/home}.

\section{RELATED WORK}

Path planning approaches for autonomous vehicles typically fall into three categories: search-based, optimization-based, and learning-based methods. Search-based techniques, such as A$^\star$\cite{a_star_orig} and its variants~\cite{a_star_2}, generate coarse obstacle-free paths, which serve as a guide for a low-level trajectory planner. Optimization-based methods incorporate detailed vehicle dynamics, optimizing trajectories for precision, comfort, and safety, but often result in computationally expensive and nonlinear NP-hard problems. Learning-based methods typically employ black-box models that map sensor inputs (e.g., camera or LiDAR data) to control actions or intermediate trajectories, often lacking interpretability.

\subsubsection{Graph search-based methods}

Algorithms such as conventional A$^\star$~\cite{lavalle2006planning} 
work only for holonomic robots since the motion primitives are linear (horizontal, vertical, diagonal). Rapidly-exploring Random Trees~(RRT)~\cite{rrt_orig} and its variants~\cite{rrt_inf, rrt_lqr} are a set of search techniques that randomly sample points within the grid-world and connect them to the tree used for planning. While such stochastic methods have greater potential to not get stuck at local optimum, they often produce paths with sharp curvature, making them difficult for the low-level trajectory follower to track. Hybrid A$^\star$~\cite{hybrid_A_star} is a search algorithm that improves upon normal A$^\star$ for non-holonomic robots. In Hybrid A$^\star$, motion between nodes follow the bicycle dynamics~\cite{rajamani2011vehicle} with a given time discretization and a fixed set of motion primitives based on speed, gear, and steering angle. Therefore, the path generated by Hybrid A$^\star$ is dynamically feasible and makes it easier for the low-level trajectory planner to track. Our work builds upon conventional Hybrid A$^\star$ and extends it to avoid dynamic obstacles using their motion predictions.  

\subsubsection{Optimization-based methods}

Advances in computational power and numerical optimization have popularized optimization-based planning methods such as Model Predictive Control~\cite{MPC}. However, obstacle avoidance often results in non-convex problems~\cite{static_parking, time_optimal_MPC, rcms}, sometimes requiring integer variables~\cite{mixed_integer_ad, slas}, making real-time implementation challenging. Prior work~\cite{OBCA, OBCA_dynamic_park} use dual variables~\cite{boyd2004convex} to smoothen constraints, enabling gradient and Hessian-based solvers. Other approaches~\cite{rapid_it, recurr_spline} leverage differential flatness~\cite{diff_flat} of the kinematic bicycle model~\cite{ rajamani2011vehicle} to generate spline-based trajectories, but still face computational complexity due to iterative optimization and curvature constraints. Hybrid methods~\cite{hybrid1, hybrid2} combine search and optimization to handle dynamic obstacles, but are not validated in tightly constrained environments such as parking lots. A time-dependent Hybrid A star algorithm was also proposed in~\cite{time_hybrid_A}, which, unlike our approach, does not exploit closed-form Reeds-Shepp path solutions, relies on a precomputed free-space representation, and employs conservative Voronoi-based collision checking. Overall, the core challenge remains the non-convexity and computational burden of optimization-based planning in dynamic settings, making real-time planning difficult.


\subsubsection{Learning-based methods}

The rich class of deep neural network models are leveraged to learn driving policies using large data from simulations or expert demonstrations. Companies such as Tesla~\cite{tesla_progress, tesla_limit} and Waymo~\cite{waymo_imit, waymo_rl} have made significant progress in autonomous driving by using learning-based methods like deep reinforcement learning or imitation learning. The ability of deep models to handle complex, unstructured environments has been explored for parking maneuvers that utilize techniques such as supervised policy learning~\cite{rl_expert} or hybrid methods combining learning with optimization techniques~\cite{hope_rl}. Despite these advancements, learning-based approaches still face challenges in generalization, safety guarantees, and real-time deployment, particularly in highly constrained scenarios with dynamic obstacles~\cite{waymo_circ}.

\section{TECHNICAL BACKGROUND}

Let $\vect{x} = (X, Y, \theta)$ be the state of the vehicle, where $(X, Y)$ is the center of the rear axis and $\theta$ is the heading angle. To model vehicle dynamics, we employ the kinematic bicycle model which is well suited for vehicle at low speeds~\cite{rajamani2011vehicle}, expressed as
\begin{equation}
\dot{\vect{x}} = f(\vect{x}, \vect{u}) \Leftrightarrow \begin{bmatrix} \dot{X} \\ \dot{Y} \\ \dot{\theta} \end{bmatrix} = \begin{bmatrix}
v \cos(\theta) \\ v \sin(\theta) \\ \frac{v}{L}\tan(\delta)
\end{bmatrix} .
    \label{dynamics}
\end{equation}
The control input is $\vect{u} = \begin{bmatrix}
    v \\ \delta
\end{bmatrix}$, where $v$ and $\delta$ is the longitudinal velocity and steering angle of the front wheel, respectively. The wheelbase of the vehicle is $L$. A path ${\mathcal{P}:=\vect{x}(t)}$, defined as the state trajectory ${\vect{x} : \mathbb{R}_{\geq 0} \to \mathbb{R}^3}$, is said to be \textit{dynamically feasible} if there exists control inputs ${\vect{u}(t) \in \mathbb{R}^2}$ for all time $t \geq 0$ such that ${\vect{x}(t) = \int_0^t f(\vect{x}(s), \vect{u}(s)) ds}$ for a given initial condition $\vect{x}(0)$.

\begin{figure}[!t]
         \centering         \includegraphics[width=0.9\linewidth]{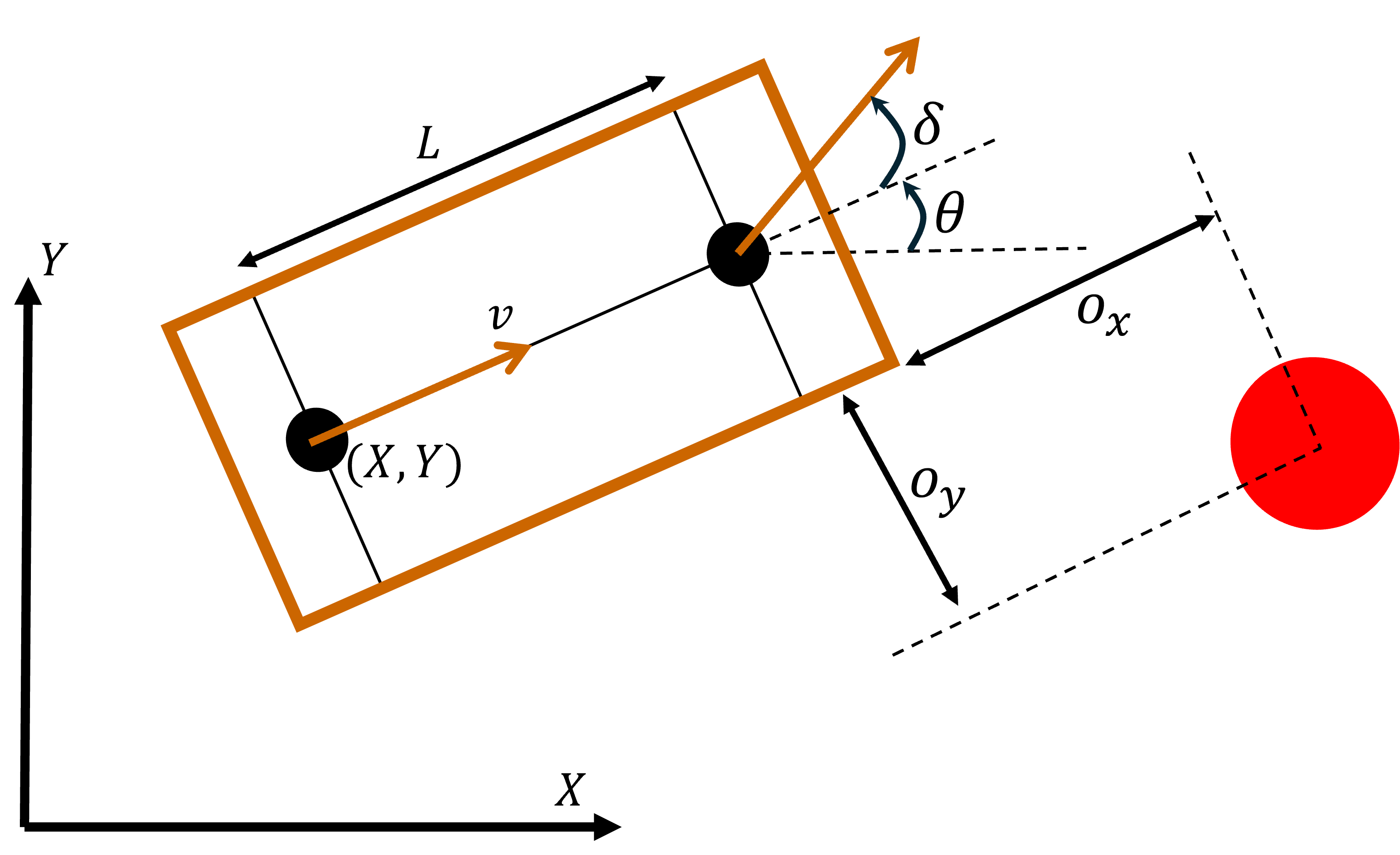}
      \caption{Geometry of the vehicle and obstacle avoidance. The vehicle is the brown rectangle and obstacle is the red circle.} 
       \vspace{-5pt}
        \label{fig:vehicle}
\end{figure}

The vehicle is modeled as a rectangle, and each obstacle is modeled as a 2D Cartesian point as given in Fig.~\ref{fig:vehicle}. Let $o_x$ be the distance from the edge of the vehicle to the obstacle along the vehicle's longitudinal direction, and let $o_y$ be the distance along the lateral direction. The obstacle avoidance constraint is
\begin{equation}
\max(o_x, o_y) \geq d,
    \label{obst_avoid}
\end{equation}
where $d > 0$ is a safety margin that intuitively enlarges the actual size of the vehicle. Constraint~\eqref{obst_avoid} precisely determines the proximity of each obstacle to the vehicle's edge, in contrast to methods that approximate obstacle distance using Euclidean distance~\cite{CBF_circle, park_circle}. Prior work also assume the obstacle to be either a polytope~\cite{OBCA_dynamic_park, OBCA} or a circle~\cite{CBF_circle, park_circle}, but we do not assume any specific shape for the obstacle. Each point on the boundary of any arbitrary shaped obstacle can be represented as the red point in Fig.~\ref{fig:vehicle}, which aligns with the raw point cloud data that we typically receive from sensors to detect obstacles~\cite{nvidia_obst_detect, obst_detect}.       

The boundaries of static obstacles are represented as a sequence of 2D Cartesian points ${\mathcal{B} = \{\vect{b}_i\}_{i=1}^{B}}$, where ${\vect{b}_i \in \mathbb{R}^2}$ and $B$ is the number of points used to model the static obstacles. For example, consider the scenario in Fig.~\ref{fig:prob_state_1}, where the ego vehicle should move from the blue state to the green state while avoiding the dynamic obstacle in red, and the static vehicles represented as black rectangles. The set $\mathcal{B}$ consists of linearly spaced points along the edges of each static vehicle, in addition to the boundaries of the drivable area. The trajectory of the $i^{\textnormal{th}}$ dynamic obstacle is given by the output of a prediction model as $\vect{y}^i(t) \in \mathbb{R}^2$ for all $i \in \{1,2,\ldots,O\}$ and time $t \geq 0$ where $O$ is the number of dynamic obstacles. In Fig.~\ref{fig:prob_state_1}, the dynamic obstacle moves horizontally from right to left at a constant velocity. Given the 2D Cartesian points for both the static and dynamic obstacles, the obstacle avoidance constraint is given by~\eqref{obst_avoid}.

\section{PROBLEM STATEMENT}

In this section, we formally define two problem statements to address the autonomous parking problem. 

\begin{problem}
\label{prob_1}
Given the initial state of the ego vehicle $\vect{x}_0$ and the goal state $\vect{e}$, a map of static obstacles $\mathcal{B}$, the predictions of dynamic obstacles $\{\vect{y}^i(s)\}_{i=1}^O$ for all $s \geq 0$, find a path $\mathcal{P} := \vect{x}(s)$ such that $\vect{x}(0) = \vect{x}_0$ and $\vect{x}(s) = \vect{e}$ for all $s \geq S$ where $S \geq 0$ is some finite time, $\mathcal{P}$ is dynamically feasible, and avoids all the static and dynamic obstacles. 
\end{problem}

\begin{figure}[!b]
         \centering         \includegraphics[width=0.4\textwidth]{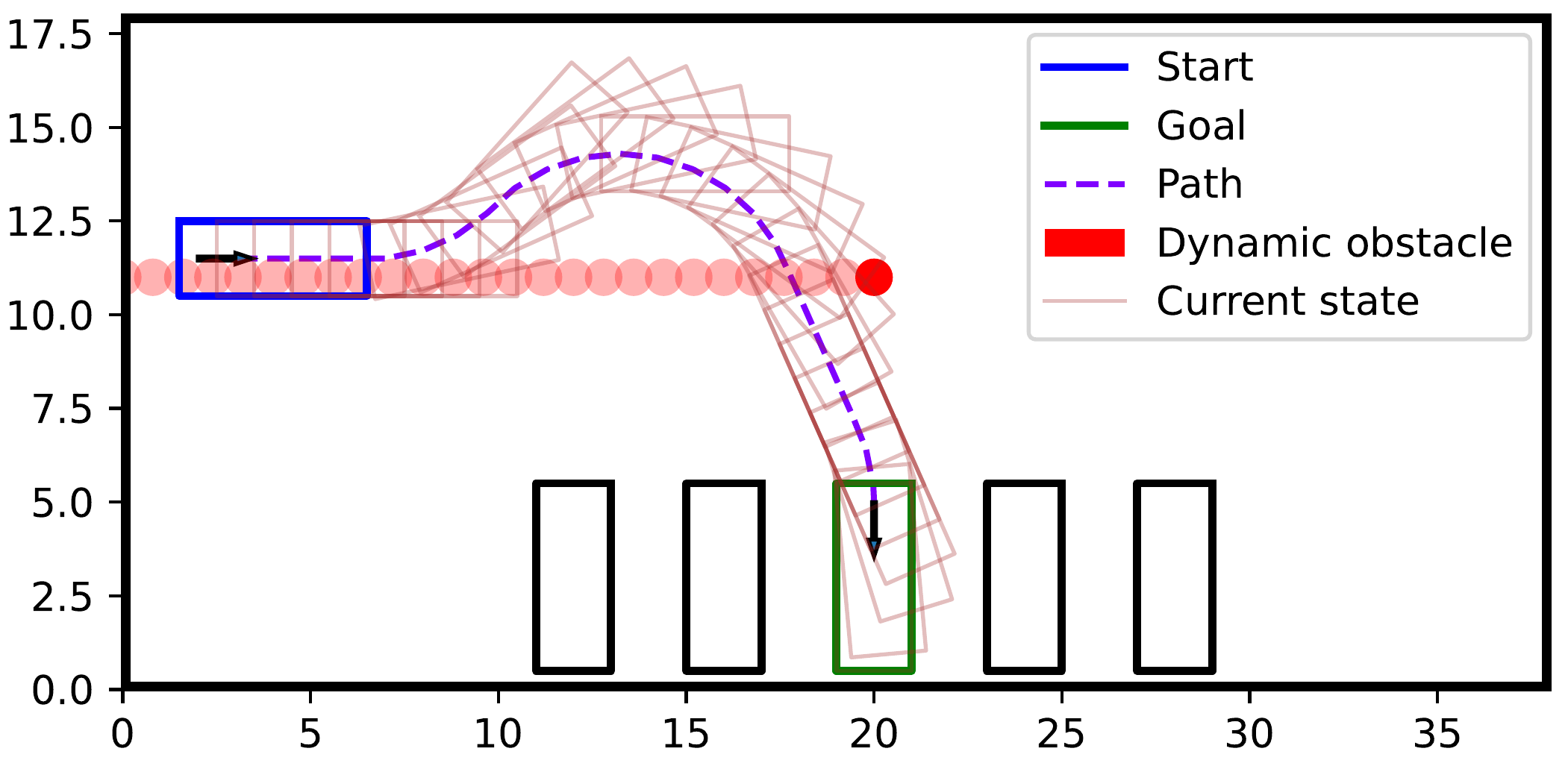}
      \caption{Target scenario for Problem~\ref{prob_1}. The black rectangles are static vehicles, and the dynamic obstacle moves from right to left.} 
       \vspace{-10pt}
        \label{fig:prob_state_1}
\end{figure}

In Problem~\ref{prob_1}, the aim is to generate a \textit{single path} from start to goal that avoids both the static and dynamic obstacles. Fig.~\ref{fig:prob_state_1} illustrates a target scenario for Problem~\ref{prob_1}. 

\begin{problem}
\label{prob_2} Given the initial state of the ego vehicle $\vect{x}_0$ and the global goal state $\vect{g}$, a map of static obstacles $\mathcal{B}$, the current state of ego vehicle $\vect{x}_t$ at time $t$, the predictions of dynamic obstacles $\vect{y}^i(s)$ for all $i \in \{1,2,\ldots,O\}$ and $s \geq 0$, find a local path $\mathcal{P}_t := \vect{x}^t(s)$ at each time $t$ such that ${\vect{x}^t(0) = \vect{x}_t}$, $\vect{x}^t(s) = \vect{g}$ for all $t \geq T, s \geq S$ where $S, T \geq 0$ is some finite time,  $\mathcal{P}_t$ is dynamically feasible, and avoids all the static and dynamic obstacles.
\end{problem}

\begin{figure}[!b]
\vspace{-3pt}
         \centering         
         \includegraphics[width=0.49\textwidth]{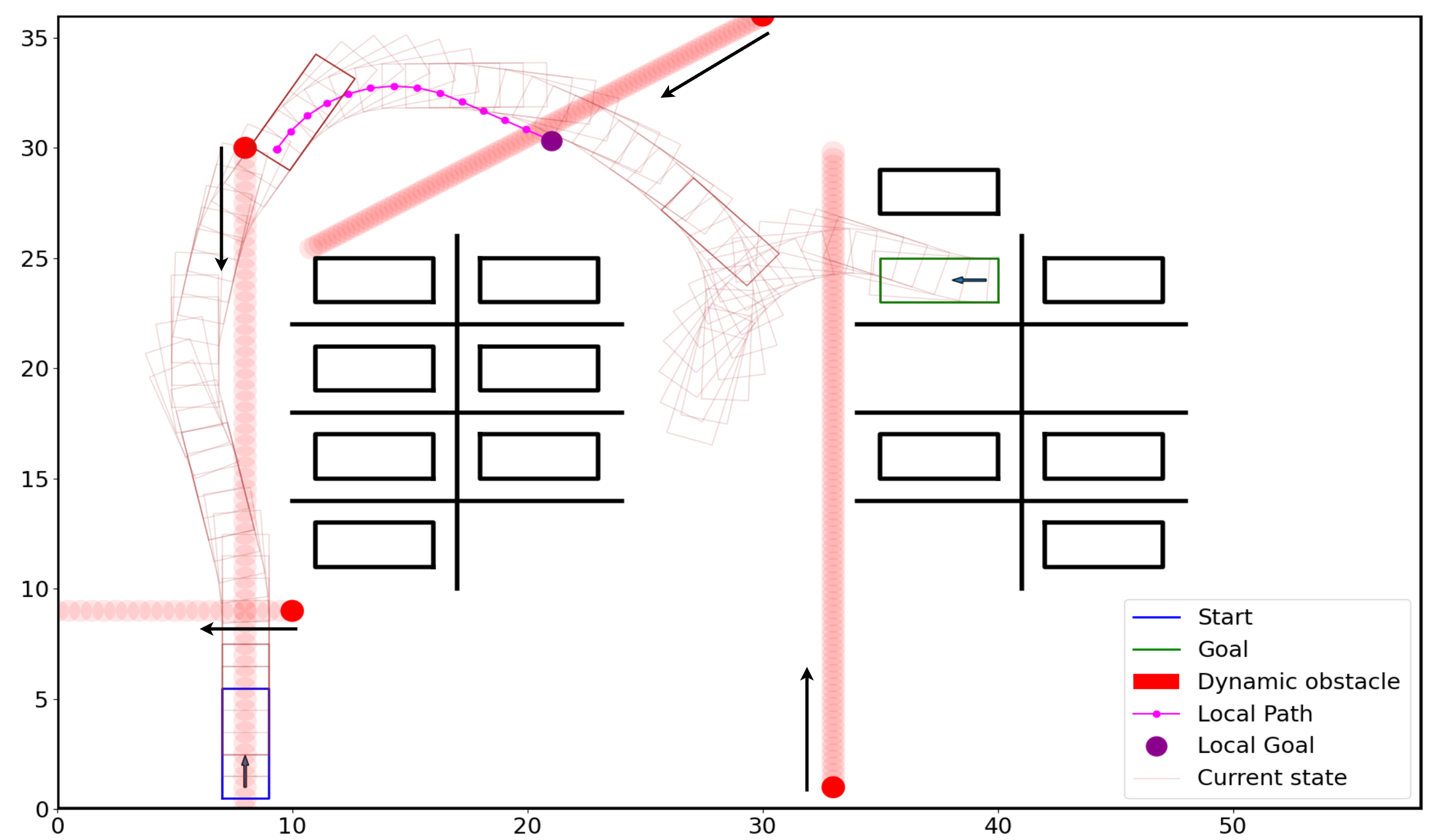}
      \caption{Target scenario for Problem~\ref{prob_2} with four dynamic obstacles where the local path of an intermediate time step is shown. The light brown rectangles denote the trajectory of the ego vehicle, and the black arrows denote the motion of dynamic obstacles.}
        \label{fig:prob_state_2}
\end{figure}

In Problem~\ref{prob_2}, the aim is to generate a local path at each time step $t$ from the current state of the vehicle $\vect{x}_t$ such that the vehicle eventually reaches the goal $\vect{g}$ without colliding with the static and dynamic obstacles. We refer to Problem~\ref{prob_2} as an \textit{online planning} problem where the vehicle should maneuver in a large parking lot as given in Fig.~\ref{fig:prob_state_2} by planning locally at each time step. Problem~\ref{prob_1} is viewed as an \textit{one-time planning} problem either for the final parking maneuver as given in Fig.~\ref{fig:prob_state_1} or to a local goal as shown in Fig.~\ref{fig:prob_state_2}. 

\begin{remark} 
\label{remark_1}
\textnormal{
In principle, for the given $\vect{x}_0$ in Problem~\ref{prob_2}, any solution that solves Problem~\ref{prob_1} with $\vect{e} = \vect{g}$ also solves Problem~\ref{prob_2} with $T = 0$. However, finding a single path $\mathcal{P}_0$ from $\vect{x}(0)$ to $\vect{g}$ in a large parking lot such as Fig.~\ref{fig:prob_state_2} will be computationally expensive. Additionally, the vehicle typically has access only to the predicted trajectories of dynamic obstacles within its local sensing region. Predictions for obstacles located very far away or for very long time horizons are often highly uncertain. Hence, we decouple the\textit{ one-time planning} problem, and the \textit{online planning} problem, referring to them as Problem~\ref{prob_1} and Problem~\ref{prob_2}, respectively. 
}
\end{remark}

\vspace{-10 pt}
\section{TIME-INDEXED HYBRID A STAR}
\vspace{-5 pt}
In this section, we propose Algorithm~\ref{alg_1} that aims to solve Problem~\ref{prob_1}. Algorithm~\ref{alg_1} is a variant of the conventional Hybrid A$^\star$ algorithm~\cite{hybrid_A_star}  where we index each node in the search procedure by time~$t$ in addition to the state~$\vect{x}$ of the vehicle. The additional time dimension allows us to account for the behavior of dynamic obstacles and subsequently check for collision during the search procedure. The details of Algorithm~\ref{alg_1} are given as follows.

\LinesNumbered
\setlength{\textfloatsep}{0pt}
\begin{algorithm}[!b]
\caption{Time-indexed Hybrid A star}
\label{alg_1}
\textbf{Input}: $\vect{x}_0, \vect{e}, \mathcal{B}, \vect{y}^i(s)$ for all $i \in \{1,2,\ldots,O\}$ and $s \geq 0$, maximum iteration $I_m$\;
\textbf{Output}: Path $\mathcal{P}$ from $\vect{x}_0$ to $\vect{e}$\;
\textbf{Initialize:} Start node $\vect{N}$, Goal node $\vect{e_N}$, \; iteration $j = 0, \mathcal{P} \gets [\vect{x}_0]$\;
\textsc{ClosedSet} = $\{\}$, \textsc{CostQueue}$[\vect{N} ] = \vect{N}.c_g + \vect{N}.c_h$\;
\While{\textnormal{goal not reached (or)} $j < I_m$}{
$j \gets j + 1$\;
$\vect{N}  \gets$ Pop \textsc{CostQueue}\;
Add $\vect{N} $ to \textsc{ClosedSet}\;
  \If{$\vect{N.x} = \vect{e}$}{
  $\mathcal{P} \gets$ \textbf{backtrack}(\textsc{ClosedSet}, $\vect{e_N}$)\;
  \Return{$\mathcal{P}$}\;
    }
    \If{$\vect{N.}h < h_{\textnormal{thresh}}$}
{$\mathcal{R} \gets$ \textbf{Reeds-Shepp}($\vect{N.x}, \vect{e}$)\;
\If{$\mathcal{R}$ \ \textnormal{is obstacle free}}
{$\mathcal{P} \gets $\textbf{backtrack}(\textsc{ClosedSet}, $\vect{e_N}$)\;
\Return{$\mathcal{P}$}\;}
}
\For{$\vect{N}' \ \textnormal{in}$ \textbf{Neighbors}$(\vect{N})$}
{
\If{$\vect{N}' \notin$ \textsc{ClosedSet} \textnormal{(and)} \big($\vect{N} ' \notin \textsc{CostQueue}$ \ \textnormal{(or)} \ $\textsc{CostQueue}[\vect{N} '] > \vect{N}'.c_g + \vect{N}'.c_h$\big)}{$\textsc{CostQueue}[\vect{N}'] = \vect{N}'.c_g+ \vect{N}'.c_h$}
}
}
\end{algorithm}

\begin{definition}
    A node~$\vect{N}$ is defined as an object with the following attributes.
    \begin{itemize}
    \item $\vect{N.x}$: state $\vect{x} = (X, Y, \theta)$ of the vehicle in node $\vect{N}$
    \item $\vect{N.}t$: time~$t \geq 0$ at node~$\vect{N}$
    \item $\vect{N.P}$: parent node of~$\vect{N}$, where $\vect{N.P}$ is the immediate predecessor of~$\vect{N}$ in the graph traversal.
    \item $\vect{N.}\tau$: state trajectory from the parent node~$\vect{N.P}$ to the current node~$\vect{N}$ for a time horizon~$H$ where ${\vect{N.}\tau(0) = \vect{N.P.}\tau(H)}$ and ${\vect{N.}\tau(H) = \vect{x}}$.
    \item $\vect{N.}\tau_o$: trajectory of the dynamic obstacles from the parent node~$\vect{N.P}$ to the current node~$\vect{N}$ for a time horizon~$H$ where ${\vect{N.}\tau_o(0) = \vect{N.P.}\tau_o(H)}$ and ${\vect{N.}\tau_o(H) = \{\vect{y}^i(\vect{N}.t)\}_{i=1}^O}$.
    \item $\vect{N}.c_g$: cost from the start node to node~$\vect{N}$
    \item $\vect{N}.c_h$: heuristic cost from node~$\vect{N}$ to the goal node.
\end{itemize}
\label{defn:node}
\end{definition}
The node~$\vect{N}$ is the fundamental entity in Algorithm~\ref{alg_1} that enables us to plan a collision-free path. It serves the same purpose as a typical node in graph-based planning algorithms~\cite{hybrid_A_star, survey_path_plan}, but we have an additonal time dimension $\vect{N.}t$ and the predictions of dynamic obstacles $\vect{N.}\tau_o$. Definition~\ref{defn:node} describes that the state of the vehicle at node~$\vect{N}$ is $\vect{N.x}$ at time~$\vect{N.}t$. The vehicle's trajectory to reach $\vect{N.x}$ from its parent node~$\vect{N.P}$ is $\vect{N}.\tau$ and the corresponding trajectory of dynamic obstacles is $\vect{N.}\tau_o$. 

\textbf{Description of Algorithm~\ref{alg_1}}: We initialize a \textsc{closedset} that contains all the explored nodes and a \textsc{costqueue} that stores the nodes to be explored sorted by the cost function $\vect{N}.c_g + \vect{N}.c_h$. In line~8, we ``pop" node~$\vect{N}$ from the \textsc{costqueue} that has the least cost. If we are ``close'' to the goal as measured using the heuristic cost $\vect{N}.c_h$ in line~14, we check if there is a direct obstacle free path to the goal using \textbf{Reeds-Shepp} paths~\cite{reeds_shepp}. If there is no such path, we explore the neighbors of the current node and add it to the \textsc{costqueue} as given in lines~22-24. We repeat the procedure until we reach the goal, or if the iteration~$j$ to explore nodes has reached $I_m$. Once we reach the goal, we \textbf{backtrack} the path by retracing the nodes from the goal to the start by following the parent nodes stored during the search. The maximum iteration~$I_m$ sets an upper limit on the computation time allocated for searching a path. If $j=I_m$, the path~$\mathcal{P}$ defaults to keeping the vehicle stationary as initialized in line~3. This will be further clarified in Section~\ref{sec:global}. 

\subsection{Neighbors}

In lines~21-23 of Algorithm~\ref{alg_1}, we explore neighboring nodes from the current node~$\vect{N}$ to find a path to the goal. The bicycle model~\eqref{dynamics} is used to generate the neighbors with a discrete set of velocity inputs $v \in [-v_{\textrm{max}}, v_{\textrm{max}}]$ and steering inputs $\delta \in [-\delta_{\textrm{max}}, \delta_{\textrm{max}}]$. A neighbor~$\vect{N'}$ to the current node~$\vect{N}$ for an input $\vect{u} =\begin{bmatrix}
    v \\ \delta
\end{bmatrix}$ and time horizon~$H$ is obtained as follows.
\begin{equation}
\begin{aligned}
\vect{N'}.\tau(r) &= \int_{0}^r f(\vect{x}(s), \vect{u}) dt \ \forall \ r \in [0, H], \vect{x}(0) = \vect{N.x} \\
\vect{N'}.\tau_o(r) &= \{\vect{y}^i(\vect{N}.t +r)\}_{i=1}^O \ \forall \ r\in[0, H] , \\
\vect{N'.x} &= \vect{N'}.\tau(H), \ \vect{N'}.t = \vect{N}.t + H, \ \vect{N'.P} = \vect{N}.   
\end{aligned}
    \label{neighbor_node}
\end{equation}
A neighboring node~$\vect{N'}$ is valid if and only if the state trajectory $\vect{N'}.\tau$ does not collide with the dynamic obstacles' trajectory $\vect{N'}.\tau_o$ and the static obstacles~$\mathcal{B}$. In practice, we discretize the trajectories and the roll out of dynamics in~\eqref{neighbor_node} at times~$\{r_1,  r_2,\ldots, r_K\}$ where $r_1 = 0$ and $r_K = H$. Collision with dynamic obstacles is checked for each $k\in\{1,2,\ldots,K\}$ by evaluating the pair of points $\left(\vect{N'}.\tau(r_k), \vect{N'}.\tau_o(r_k\right)$ using~\eqref{obst_avoid}. 

\subsection{Reeds-Shepp}

If the current node~$\vect{N}$ is ``close" to the goal, we check if there exists a continuous obstacle free path to the goal. The current node is``close" to the goal if the heuristic cost function $\vect{N}.c_h$ is less than a threshold $h_{\textnormal{thresh}}$. We compute all possible Reeds-Sheep paths~\cite{reeds_shepp} from the current state $\vect{N.x}$ to the goal state~$\vect{e}$. A Reeds-Shepp path is the optimal path between two states for a vehicle with bicycle dynamics~\eqref{dynamics} that moves only forward~${(v = v_{\textnormal{max}})}$ or backward~${(v = -v_{\textnormal{max}})}$ with extreme steering inputs~${\delta \in \{-\delta_{\textrm{max}}, \delta_{\textrm{max}}\}}$. The obstacle free Reeds-Shepp Paths~$\mathcal{R}$ are sorted as per the user-defined cost~$c_g$ and the path with the least cost is chosen to move from the current state to the goal state.

\subsection{Cost}

The node $\vect{N}$ in line~8 has the least cost $\vect{N}.c_g + \vect{N}.c_h$ amongst all the nodes in \textsc{costqueue}. The cost function $\vect{N}.c_g$ penalizes path length, steering angle, reverse motion and change in direction of motion and steering between subsequent nodes. The heuristic cost $\vect{N}.c_h$ guides the search direction towards the goal using an under-estimate of the actual cost to the goal $\vect{e_N}.g$. We use the solution of the $A^\star$ algorithm~\cite{a_star_orig} for the heuristic cost which computes the shortest paths from each discrete point in a grid-world environment to the goal without using bicycle dynamics. We pre-compute the $A^\star$ costs using only the static obstacles, since dynamic obstacle avoidance is handled by our time-indexed Hybrid A$^\star$ algorithm.

As described in Remark~\ref{remark_1}, computing a single path from start to the global goal for a scenario such as Fig.~\ref{fig:prob_state_2} will not be feasible and practical. In the next section, we present an online planning strategy for parking in larger lots. 

\section{ONLINE PLANNER}
\label{sec:global}
 
In this section, we propose Algorithm~\ref{alg_2} that aims to solve Problem~\ref{prob_2}, where we find a local path~$\mathcal{P}_t$ at each online planning step~$t$ that moves the vehicle towards the goal while avoiding dynamic obstacles. 

We first initialize a global path~$\mathcal{G}$ from $\vect{x}_0$ to $\vect{g}$ using conventional Hybrid A$^\star$~\cite{hybrid_A_star} that avoids the static obstacles. Then, at each online planning step~$t$, given the current state~$\vect{x}_t$ and the predictions of dynamic obstacles~$\vect{y}^i(s)$ for all ${i\in\{1,2,\ldots,O\}}$ and $s \geq 0$, we compute a local path~$\mathcal{P}_t$ that follows the global path~$\mathcal{G}$ while avoiding dynamic obstacles. We utilize Algorithm~\ref{alg_1} as a sub-routine in Algorithm~\ref{alg_2} to plan the local path. The key idea in Algorithm~\ref{alg_2} is to choose an appropriate intermediate goal~$\vect{e}$ for Algorithm~\ref{alg_1} to return a feasible local path~$\mathcal{P}_t$. 

\LinesNumbered
\begin{algorithm}[!b]
\caption{Online Planner}
\label{alg_2}
\SetAlgoLined
\textbf{Given}: $\vect{x}_0, \vect{g}, \mathcal{B}$, maximum look-ahead=$N_m$, maximum iteration =$I_m$\;
\textbf{Initialize:} $\mathcal{G} \gets $ global path from $\vect{x}_0$ to $\vect{g}$ using vanilla Hybrid A$^\star$ that avoids $\mathcal{B}$\;
\textbf{Input}: $\vect{x}_t, \vect{y}^i(s)$ for all $i \in \{1,2,\ldots,O\}$\;
\textbf{Output}: Local path $\mathcal{P}_t$\;
\textbf{Initialize:}
$\mathcal{P}_t \gets [\vect{x}_t]$\;
$i_c \gets \textrm{argmin}_{i \in \{1,2,\ldots,|\mathcal{G}|\}} \|\vect{x}_t - \mathcal{G}[i]\|$\;
$i_g \gets i_c + N_m$\;
\While{$\mathcal{P}_t = \vect{x}_t$ \ \textnormal{(and)} $i_g > i_c$}
{
$\vect{e} \gets \mathcal{G}[i_g]$\;
$\mathcal{P}_t \gets$ \hyperref[alg_1]{Algorithm1}$\left(\vect{x}_t,\vect{e}, \mathcal{B}, \{\vect{y}^i(s)\}_{i=1}^O, I_m\right)$\;
$i_g \gets i_g - 1$\;
}
 \end{algorithm}
 
\textbf{Choosing intermediate goal}: At the current planning step, we find the closest point on the global path~$\mathcal{G}$ to the current state~$\vect{x}_t$. Then, we initialize the intermediate goal using a look-ahead $N_m$ from the closest point on $\mathcal{G}$. In line~10, we try to compute a path from $\vect{x}_t$ to the intermediate goal~$\vect{e}$ that avoids all the obstacles within a finite time, which is directly proportional to the runtime of Algorithm~\ref{alg_1}. Instead of the actual runtime, we use the maximum iteration~$I_m$ of Algorithm~\ref{alg_1} as the stopping criteria. If Algorithm~\ref{alg_1} does not find a path within iteration~$I_m$, then we adapt the intermediate goal to be one index closer to $\vect{x}_t$ along~$\mathcal{G}$ and run Algorithm~\ref{alg_1} again. We repeatedly adapt the intermediate goal as given in lines~$9-11$ until we find a path~$\mathcal{P}_t$. If the intermediate goal is the closest point on $\mathcal{G}$ to $\vect{x}_t$, we return the local path to be $[\vect{x}_t]$ as initialized in line~5 of Algorithm~\ref{alg_2}, which commands the vehicle to stay stationary. This intuitively explains that if the planner cannot find a feasible local path that follows the global path within a reasonable time, the vehicle stays stationary. 
 
\section{EXPERIMENTS}

We evaluate our proposed Algorithms on a set of common parking scenarios, assuming that the perception system provides real-time information on static vehicles, curb boundaries, and dynamic pedestrians within the system’s perception range. The task for the ego vehicle is to navigate from the initial position to a goal state while avoiding collisions. The vehicle dynamics are described by the kinematic bicycle model~\eqref{dynamics} discretized with a time step of $0.1$ s for our simulations. The vehicle parameters are given in Table~\ref{table:veh_params}. We use Honda Odyssey~\cite{honda_od} as a reference to set the dimensions of the vehicle. We use $1$ m/s as the velocity limit in our Algorithms to encourage caution, but will increase this limit in our future work. All simulations are performed in Python 3.8 on Ubuntu 20.04 with Intel Xeon E5-2643 v4 CPU. 

Our time-indexed Hybrid A$^\star$ implementation uses a state grid size of 2 m in both $X$ and $Y$ directions, and $20$ deg in the heading angle. The steering input is discretized using 5 points in $[-\delta_{\textrm{max}}, \delta_{\textrm{max}}]$ and velocity is discretized using 3 points in $[-v_{\textrm{max}}, v_{\textrm{max}}]$. The length of each parking space is $6.5$ m, and the width is $3.5$ m as referred from ~\cite{park_dim}. The inclination angle with respect to the driving direction is $70$~deg for angle parking. Each static vehicle is modeled as a rectangle with length and width given in Table~\ref{table:veh_params}, and each dynamic obstacle is modeled as a circle of radius $0.5$~m. Collision is checked using the geometry of the vehicle in Fig.~\ref{fig:vehicle} with a safety threshold of $d = 0.5$~m.  

\bgroup
\def\arraystretch{1.2}
\centering
\begin{table}[!b]
 \captionsetup{justification=centering}
\caption{Vehicle parameters}
\centering
\begin{tabular}{||P{1.2 cm}|P{2.1cm}|P{1.2cm}||}
            \hline
            \textbf{Parameter} & \textbf{Description} & \textbf{Value} \\
            \hline 
            \hline
            $L$ & Wheelbase length & 3 \ m \\
            \hline
            $V_L$ & Vehicle length & 5 \ m \\
            \hline
            $V_W$ & Vehicle width & 2 \ m \\
            \hline
            $v_{\textnormal{max}}$ & Velocity limit & 1 \ m s$^{-1}$ \\
            \hline
            $\delta_{\textnormal{max}}$ & Steering limit & 40 \ deg  \\
            \hline
\end{tabular}
\label{table:veh_params}
\end{table}
\egroup

\vspace{-5pt}

\begin{figure*}[!b]
\vspace{-15pt}
         \centering         
         \includegraphics[width=\textwidth]{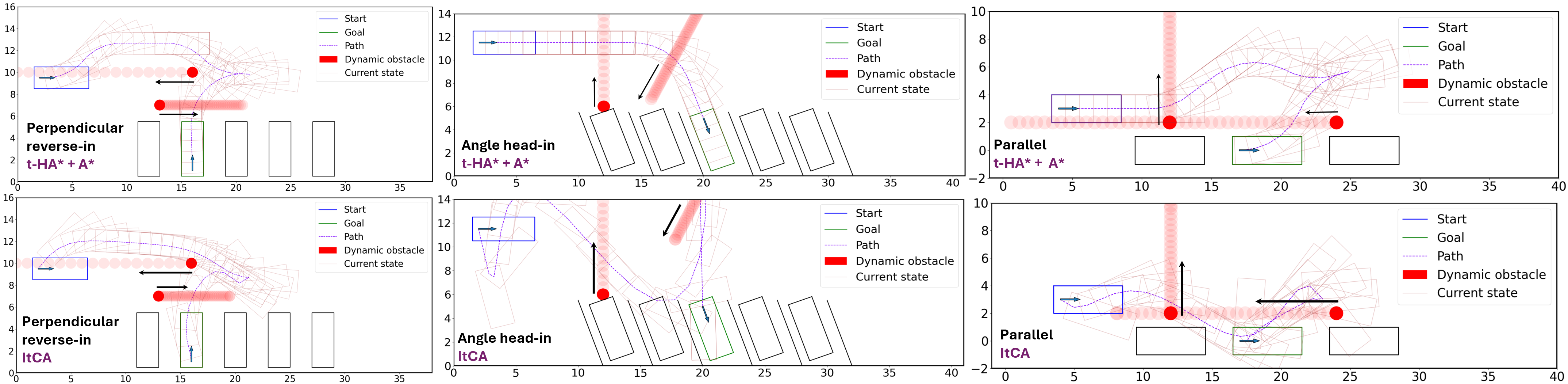}
      \caption{\small Comparison of paths generated by our time-indexed Hybrid A* method~(t-HA* + A*) and the iterative spline-based method~(ItCA).}
       \vspace{-10pt}
        \label{fig:spline_ours}
\end{figure*}

\begin{figure*}[!b]
         \centering         
         \includegraphics[width=\textwidth]{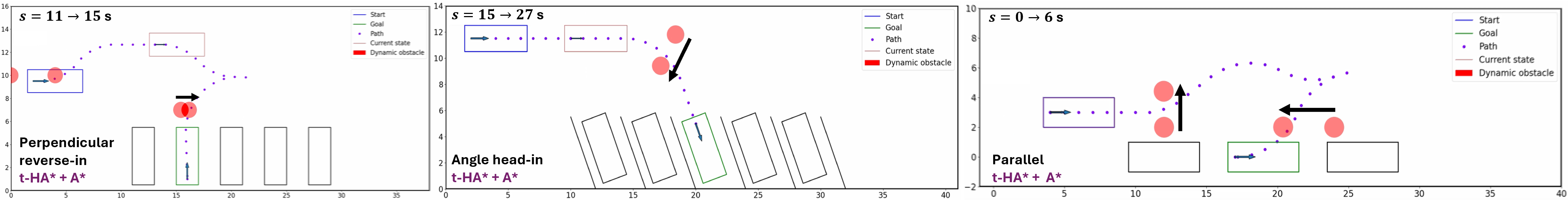}
      \caption{\small Illustration of the ego vehicle's stationary behaviors at different time steps~$s$ in the paths generated by Algorithm~\ref{alg_1}~(t-HA* + A*).}
        \label{fig:snaps_stop_go}
\end{figure*}

\subsection{One-time Planning}
\label{sec:exp_local}

We validate Algorithm~\ref{alg_1} with $I_m=500$ for the following parking scenarios as given in Fig.~\ref{fig:spline_ours}: perpendicular head-in, perpendicular reverse-in, angle head-in and parallel. In all the cases presented in this work, the ego vehicle parks in confined spaces with defined outer boundaries in the presence of other parked vehicles and dynamic obstacles. We compare Algorithm~\ref{alg_1} --- the time-indexed Hybrid A* (\textbf{t-HA*}) --- with the state of the art iterative spline-based collision avoidance method~\cite{rapid_it} (ItCA). The work in~\cite{rapid_it} iteratively refines $5^{\textrm{th}}$-order spline trajectories to track a reference path (e.g., Hybrid A* avoiding static obstacles). If collisions with (possibly dynamic) obstacles occur, the tracking cost is relaxed at collision points in each iteration, until the path is collision-free, or a maximum iteration count is reached.

In Fig.~\ref{fig:spline_ours}, we qualitatively compare the paths generated by ItCA~\cite{rapid_it} with that of Algorithm~\ref{alg_1} using the A* heuristic (\textbf{t-HA* + A*}). The paths generated by {t-HA* + A* avoids both static and dynamic obstacles in all scenarios. The ItCA method generates collision-free paths in the absence of dynamic obstacles for all the parking scenarios presented in this work. However, in the presence of multiple dynamic obstacles, the ItCA path for reverse-in parking collides with a static obstacle, and the paths for angle and parallel parking collide with obstacles at multiple points. We rigorously test our simulation runs across different initial positions and velocities of the dynamic obstacles. The initial point of each obstacle are regularly spaced within the bounds given in Table~\ref{table:park_params}. The bounds in Table~\ref{table:park_params} are chosen so that the dynamic obstacles cover a sufficient area of the drivable region and are possibly on a collision course with the ego vehicle to verify if the algorithms can avoid the obstacles. We generate 100 initial points for the one obstacle in perpendicular head-in parking~(Fig.~\ref{fig:prob_state_1}). We choose 15 points for each obstacle in perpendicular reverse-in, angled head-in, and parallel parking~(Fig.~\ref{fig:spline_ours}), resulting in a total of $15^2 = 225$ initial position pairs. The velocity of obstacles in the $X$ and $Y$ directions for each candidate initial point is sampled from the uniform distribution~$[-0.7, 0.7]$~m/s. We run 50 tests for each set of initial points and velocities. We also evaluate two variations of t-HA*, one with a Eucledian heuristic  cost function~(t-HA* + Eucledian) and a grid-based A* cost~(\textbf{t-HA* + A*}). 

The average performance and trajectory quality metrics for the experiments, along with one standard deviation, are presented in Table~\ref{table:graph_results}. The average runtime of t-HA* + A* is 10-100 times faster than ItCA or when using the Eucledian heuristic. Since the A* heuristic exploits the information of static obstacles, the search procedure in Algorithm~\ref{alg_1} explores nodes more optimally than when the Eucledian heuristic is used. The heading rate and curvature in ItCA are significantly larger than in t-HA*, rendering the spline paths infeasible for the vehicle to follow given its velocity and steering limits~(Table~\ref{table:veh_params}). The curvature is computed using the formula provided in Section~III-E of~\cite{recurr_spline}.

\bgroup
\def\arraystretch{1.4}
\centering
\begin{table}[!t]
 \captionsetup{justification=centering}
\caption{Parameters of the Parking Scenarios, where Initial and Goal state of the ego vehicle is $[X \ \textrm{m}, Y \ \textrm{m}, \theta \ \textrm{deg}]$  and $(X_i \ \textrm{m}, Y_i \ \textrm{m})$ is the Initial Position of Obstacle~$i$.}
\centering
\begin{tabular}{||P{2.2 cm}|P{1.8cm}|P{3.2 cm}||}
            \hline
            \textbf{Scenario} & \textbf{Initial, Goal State $[X, Y, \theta ]$}  & \textbf{Initial Position of Dynamic Obstacles}  \\
            \hline 
            \hline
            Perpendicular head-in~(Fig.~\ref{fig:prob_state_1}) & $[2.0, 11.5, 0]$, \newline $[20.0, 5.0, -90]$ & $X_1, Y_1\sim[15, 30], [7, 17]$ \\
            \hline
            Perpendicular reverse-in~(Fig.~\ref{fig:spline_ours}) & $[2.0, 9.5, 0]$, \newline $[16.0, 1.0, 90]$ & $X_1, Y_1\sim[12, 25], [8, 13]\newline X_2, Y_2\sim[10, 18], [5, 8]$ \\
            \hline
           Angle head-in~(Fig.~\ref{fig:spline_ours}) & $[2.0, 11.5, 0]$, \newline $[21.0, 5.0, -70]$ & $X_1, Y_1\sim[11, 16], [5, 9], \newline X_2, Y_2\sim[16, 25], [10, 15]$ \\
            \hline
            Parallel~(Fig.~\ref{fig:spline_ours}) & $[4.0, 3.0, 0]$, \newline $[17.0, 0.0, 0]$ & $X_1, Y_1\sim[10, 15], [1, 3], \newline X_2, Y_2\sim[20, 28], [2, 4]$\\
            \hline
\end{tabular}
\label{table:park_params}
\end{table}
\egroup

 In Fig.~\ref{fig:snaps_stop_go}, the paths generated by t-HA* show that the ego vehicle yields to dynamic obstacles before proceeding to the parking spot. This mirrors typical human driving behavior by pausing to assess and adapt, ensuring both safety and smooth navigation. In contrast, ItCA often produces sharp turns, leading to infeasible curvature or failure to find a collision-free path within the iteration limit. Even with an increased limit of 100 — compared to 10 used in~\cite{rapid_it} — ItCA exhibits a high failure rate as given in Table.~\ref{table:graph_results}, except when dynamic obstacles are positioned far from the initial reference path.

\begin{figure*}[!b]
        \vspace{-10pt}
         \centering         
         \includegraphics[width=\textwidth]{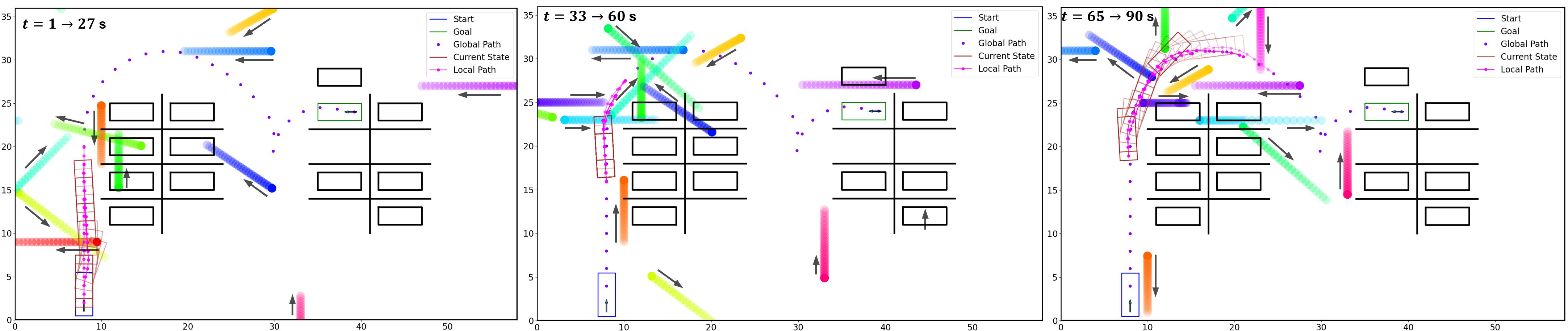}
      \caption{\small Perpendicular reverse-in parking in a large surface lot with 15 dynamic obstacles shown as different coloured circles.}
        \label{fig:lot_15}
\end{figure*}

\bgroup
\def\arraystretch{1.4}
\centering
\begin{table*}[!t]
    \caption{Comparison of Average Performance and Trajectory Metrics, with One Standard Deviation, for all the Parking Scenarios.}
 \captionsetup{justification=centering}
    \centering
    \begin{tabular}{||P{1.4 cm}|P{1.1 cm}|P{1.5 cm}|P{0.8 cm}|P{1.7 cm}|P{1.9cm}|P{1.7cm}|P{1.9cm}|P{1.7cm}||}
        \hline
        \textbf{Scenario} & \textbf{Number of dynamic obstacles} & \textbf{Method } & \textbf{Failure Rate $\downarrow$} & \textbf{Runtime } \newline  $\downarrow$ (s) & \textbf{Path length} \newline  $\downarrow$ (m) & \textbf{Distance to closest obstacle} \newline $\uparrow$ (m) & \textbf{Heading rate} \newline $\downarrow$ $\left(\textnormal{deg} \ \textnormal{s}^{-1}\right)$ & \textbf{Curvature} \newline $\downarrow$ $\left(\textnormal{m}^{-1}\right)$\\
        \hline
        \multirow{3}{1.4 cm}{Perpendicular head-in (Fig.~\ref{fig:prob_state_1})} & \multirow{3}{*}{1} & ItCA  & 8\% & $0.823 \pm 0.144$	& $44.902 \pm	8.277 $ & $2.246 \pm 0.247$ & $5.711 \pm	4.202$ &	$0.118 \pm	0.098$ \\ 
        \cline{3-9}
        & & t-HA* + Eucledian & 0.00\% & $0.448\pm 0.061$ & $43.223 \pm 6.084$& $2.314 \pm	0.261$ & 	$4.708 \pm	4.165$ & $0.074 \pm 0.071$ \\ 
        \cline{3-9}
        & & \textbf{t-HA* + A*} & \textbf{0.00\%} & \textbf{0.018 $\pm$ 0.004} & \textbf{40.578 $\pm$ 8.312} & \textbf{2.983 $\pm$ 0.476} & \textbf{3.009 $\pm$ 1.461} & \textbf{0.051 $\pm$ 0.026} \\ 
        \hline
        \multirow{3}{1.4 cm}{Perpendicular reverse-in (Fig.~\ref{fig:spline_ours})} & \multirow{3}{*}{2} & ItCA & 67.6\% & $1.485 \pm 0.84$ &	$33.624 \pm	9.216$	& $1.538 \pm 0.182$ &	$13.85 \pm	20.355$	& $1.21 \pm	9.7 $ \\ 
        \cline{3-9}
        & & t-HA* + Eucledian  & 0.00\% & $9.716 \pm	0.217$ &	$40.902 \pm	8.936$ &	$1.592 \pm	0.435$	& $4.867 \pm 5.843$ &	$0.086 \pm	0.101$ \\ 
        \cline{3-9}
        & & \textbf{t-HA* + A*}  & \textbf{0.00\%} & \textbf{0.048} $\pm$ \textbf{0.007} & \textbf{28.967 $\pm$ 4.569} & \textbf{1.75 $\pm$	0.347} & \textbf{4.401 $\pm$ 2.388} & \textbf{0.072 $\pm$ 0.036} \\ 
        \hline
        \multirow{3}{1.4 cm}{Angle head-in (Fig.~\ref{fig:spline_ours})} & \multirow{3}{*}{2} & ItCA & 85.7\% & $4.998 \pm 0.428$ &	$56.134 \pm	11.934$	& $0.616 \pm 0.241$ &	$25.67 \pm	33.327$ &	$0.628 \pm	2.294$ \\ 
        \cline{3-9}
        & & t-HA* + Eucledian & 0.00\% & $4.749 \pm	0.211$ & $34.872 \pm 10.944$ & $1.374 \pm	0.385$ & $3.826 \pm	0.917$ & $0.047 \pm	0.041$ \\ 
        \cline{3-9}
        & & \textbf{t-HA* + A*} & \textbf{0.00\%} & \textbf{0.034 $\pm$	0.005} & \textbf{27.08 $\pm$ 6.813} & \textbf{1.374 $\pm$	0.065} & \textbf{2.471 $\pm$ 0.779} & \textbf{0.042 $\pm$ 0.013} \\ 
        \hline
        \multirow{3}{1.4 cm}{Parallel (Fig.~\ref{fig:spline_ours})} & \multirow{3}{*}{2} & ItCA & 92\% &  $0.781 \pm	0.309$	& $38.924 \pm	9.869$ &	$0.953 \pm  0.176$ & $40.949 \pm	41.067$ &	$0.852 \pm	2.463$ \\ 
        \cline{3-9}
        & & t-HA* + Eucledian & 0.00\% & $6.077 \pm	0.432$ & \textbf{30.179 $\pm$ 7.557} & \textbf{2.045 $\pm$	0.336} &	\textbf{3.619 $\pm$	5.514} &	\textbf{0.058 $\pm$	0.088} \\ 
        \cline{3-9}
        & &\textbf{t-HA* + A*} & \textbf{0.00\%} & \textbf{0.026 $\pm$	0.001} & $32.713 \pm 2.481$ & $1.686 \pm	0.169$ & $6.421 \pm 0.819$ & $0.106 \pm	0.014$ \\         
        \hline
        \multirow{3}{1.4 cm}{Perpendicular reverse-in surface lot (Fig.~\ref{fig:lot_15})} & 4 & \textbf{t-HA* + A*} & 15\% & $0.012 \pm 	0.005$ &	$92.368 \pm	6.829$ & $2.058 \pm	0.191$ & $3.694 \pm	0.58$ & $0.062 \pm	0.028$ \\ 
        \cline{2-9}
        & 10 & \textbf{t-HA* + A*} & 30\% & $0.012 \pm	0.002$ & $95.353 \pm 19.461$ & 	$1.912 \pm	0.159$ &	$3.278 \pm	4.32$ &	$0.051 \pm	0.07$ \\
        \cline{2-9}
        & 15 & \textbf{t-HA* + A*} & 35\% & $0.011 \pm	0.016$ &	$76.141 \pm 11.19$	& $1.546 \pm	0.213$ & $2.804 \pm	4.314$ &	$0.043 \pm	0.07$\\
        \hline
    \end{tabular}
    \label{table:graph_results}
    \vspace{-17pt}
\end{table*}
\egroup

\vspace{-5pt}
\subsection{Online Planning}
\label{sec:exp_global}

We validate Algorithm~\ref{alg_2} for perpendicular reverse-in parking in a large surface lot depicting a dense traffic situation that includes multiple parked cars and dynamic obstacles as given in Fig.~\ref{fig:lot_15}. We use a maximum look-ahead of $N_m = 5$ to choose an intermediate goal from the initial global path~$\mathcal{G}$, and maximum iteration~$I_m = 100$ as the stopping criteria in Algorithm~\ref{alg_2}. All other vehicle parameters and parking dimensions are the same as used in Section~\ref{sec:exp_local}. The road width that separates adjacent parking rows is $10$~m.

The trajectory of the ego vehicle navigating to a designated parking spot is illustrated in Fig.~\ref{fig:lot_15} with 15 dynamic obstacles. Each local path connects the current state of the ego vehicle to an intermediate goal, selected as described in Section~\ref{sec:global}. The maximum iteration limit~$I_m$ ensures that Algorithm~\ref{alg_2} operates at a high enough frequency, enabling frequent refinement of the local path to closely track the global path while avoiding obstacles.

\bgroup
\def\arraystretch{1.1}
\centering
\begin{table}[!b]
 \captionsetup{justification=centering}
        \vspace{2pt}
\caption{Initial Position of Dynamics Obstacles for Surface Lot Environment~(Fig.~\ref{fig:lot_15}).}
        \vspace{-1pt}
\centering
\begin{tabular}{||P{2.2 cm}|P{2.2cm}|P{3 cm}||}
            \hline
            \textbf{Total number of dynamic obstacles} & \textbf{Split of dynamic obstacles}  & \textbf{Initial Position} $X \ \textrm{m}, Y \ \textrm{m}\sim$ \\
            \hline 
            \hline
            \multirow{2}{*}{4} & 2 & [0, 15], [7 , 40]\\
            \cline{2-3}
            & 2 & [25, 40],  [0, 40]\\
            \hline
            \multirow{2}{*}{10} & 6 & [0, 20], [7, 60]\\
            \cline{2-3}
            & 4 & [20, 40],  [0, 60]\\
            \hline
            \multirow{3}{*}{15} & 5 & [0, 20],  [7, 20]\\
            \cline{2-3}
            & 4 & [0, 20],  [20, 60]\\
            \cline{2-3}
            & 6 & [20, 40],  [0, 60]\\
            \hline
\end{tabular}
\label{table:lot_15_params}
        \vspace{-8pt}
\end{table}
\egroup

\vspace{-1 pt}
We run experiments for three cases in the large surface lot environment, where each case corresponds to different number of dynamic obstacles: 4, 10 and 15. We conduct 20 experiments for each of the three cases, and 10 test runs for each experiment. The initial positions of the dynamic obstacles in each experiment are randomly sampled from the uniform distributions specified in Table~\ref{table:lot_15_params}. Different bounds are used for different obstacle sets so that the ego vehicle encounters dense traffic situations at varying time instances. The velocity of obstacles in the $X$ and $Y$ directions for each candidate initial point is sampled from the uniform distribution~$[-0.7, 0.7]$ m/s. The initial state is $(8\ \textrm{m}, 1\ \textrm{m}, 90\ \textrm{deg})$ and the goal state is $(40\ \textrm{m}, 24\ \textrm{m}, 180\ \textrm{deg})$. The performance metrics for the surface lot scenarios are also summarized in Table~\ref{table:graph_results}. The relatively higher failure rate of t-HA* + A* in the surface lot environment is due to the absence of a feasible path caused by dense traffic conditions. In contrast, ItCA and t-HA* + Euclidean always generate unsafe paths, even when the iteration limit is increased to 1000, corresponding to an average maximum timeout of 10 seconds. The runtime for the surface lot scenario in~Fig.\ref{fig:lot_15} corresponds to Algorithm~\ref{alg_2}, which computes a local path at each time step, unlike other scenarios described in Section~\ref{sec:exp_local} that use Algorithm~\ref{alg_1}. The lower run-time for Fig.~\ref{fig:lot_15} is due to planning toward a nearer intermediate goal compared to the more distant goals in Figs.\ref{fig:prob_state_1} and\ref{fig:spline_ours} The trajectory metrics, including path length, distance to the closest obstacle, heading rate, and curvature, for the surface lot scenario, are calculated over the entire trajectory, from the start to the final goal state, while deviating slightly from the global path to avoid obstacles.


\section{CONCLUSION AND FUTURE WORK}

We proposed a \textit{time-indexed} Hybrid A* algorithm that explicitly incorporates dynamic obstacle predictions to generate safe, reliable, and smooth paths across diverse parking scenarios. This was further extended to an online planning strategy for large surface lots via an adaptive goal-selection mechanism. Simulations across multiple parking settings demonstrate improved computational efficiency, safety, and feasibility over the state-of-the-art spline-based planner.

Future work will address the assumption of perfect trajectory predictions by the current method for dynamic obstacles, which overlooks uncertainties and latencies in the planning pipeline. We also aim to relax the assumption of a known goal state by extending our approach to actively explore the parking lot and identify suitable parking spots, all while accounting for uncertainty in the behavior of other agents.






\vspace{-5pt}
\bibliographystyle{IEEEtran}
\bibliography{IEEEabrv, IEEEexample}

\end{document}